\documentclass[runningheads]{llncs}

\usepackage{eccv}

\usepackage{eccvabbrv}

\usepackage{graphicx}
\usepackage{booktabs}

\usepackage[accsupp]{axessibility}  

\usepackage[pagebackref,breaklinks,colorlinks,citecolor=eccvblue]{hyperref}

\usepackage{orcidlink}

\begin{document}

\title{Depth-based Privileged Information for Boosting 3D Human Pose Estimation on RGB} 

\titlerunning{Depth-based Privileged Information}

\author{
Alessandro Simoni\inst{1} 
\and
Francesco Marchetti\inst{2}
\and
Guido Borghi\inst{1}
\and
Federico Becattini\inst{3}
\and
Davide Davoli\inst{4}
\and
Lorenzo Garattoni\inst{4}
\and
Gianpiero Francesca\inst{4}
\and 
Lorenzo Seidenari\inst{2} 
\and
Roberto Vezzani\inst{1}
}

\authorrunning{A.~Simoni, F.~Marchetti et al.}

\institute{
University of Modena and Reggio Emilia, Modena, 41100, Italy\\
\email{\{a.simoni, guido.borghi, roberto.vezzani\}@unimore.it}
\and
University of Florence, Firenze, 50134, Italy\\
\email{\{francesco.marchetti, lorenzo.seidenari\}@unifi.it}
\and 
University of Siena, Siena, 53100, Italy\\
\email{federico.becattini@unisi.it}
\and
Toyota Motor Europe\\
\email{\{davide.davoli, lorenzo.garattoni, gianpiero.francesca\}@toyota-europe.com}
}

\maketitle

\begin{abstract}
Despite the recent advances in computer vision research, estimating the 3D human pose from single RGB images remains a challenging task, as multiple 3D poses can correspond to the same 2D projection on the image. 
In this context, depth data could help to disambiguate the 2D information by providing additional constraints about the distance between objects in the scene and the camera. Unfortunately, the acquisition of accurate depth data is limited to indoor spaces and usually is tied to specific depth technologies and devices, thus limiting generalization capabilities.
In this paper, we propose a method able to leverage the benefits of depth information without compromising its broader applicability and adaptability in a predominantly RGB-camera-centric landscape.
Our approach consists of a heatmap-based 3D pose estimator that, leveraging the paradigm of Privileged Information, is able to hallucinate depth information from the RGB frames given at inference time.
More precisely, depth information is used exclusively during training by enforcing our RGB-based hallucination network to learn similar features to a backbone pre-trained only on depth data.
This approach proves to be effective even when dealing with limited and small datasets.
Experimental results reveal that the paradigm of Privileged Information significantly enhances the model's performance, enabling efficient extraction of depth information by using only RGB images.
  \keywords{3D Pose Estimation \and Privileged Information \and RGB-based Hallucination}
\end{abstract}

\section{Introduction}
\label{sec:introduction}
The ability to estimate the absolute 3D pose of humans from images is a key technology for several applications, such as human-robot interaction~\cite{simoni2022semi}, activity recognition~\cite{luvizon20182d}, augmented and virtual reality~\cite{lin2010augmented}, smart factories~\cite{wu2021human}. 
However, despite the recent advances of artificial intelligence and computer vision in the area of video analysis~\cite{chen2018cascaded, sun2019deep, toshev2014deeppose}, estimating the absolute 3D human pose from monocular RGB images is a problem that has not yet been completely solved. Indeed, the direct estimation from monocular RGB images can be viewed as an ill-posed problem due to the depth ambiguity, \ie the fact that different 3D skeleton configurations could correspond to the same 2D projection on the image plane \cite{fabbri2020compressed}.


\begin{figure}[t]
    \centering
    \includegraphics[width=0.65\linewidth]{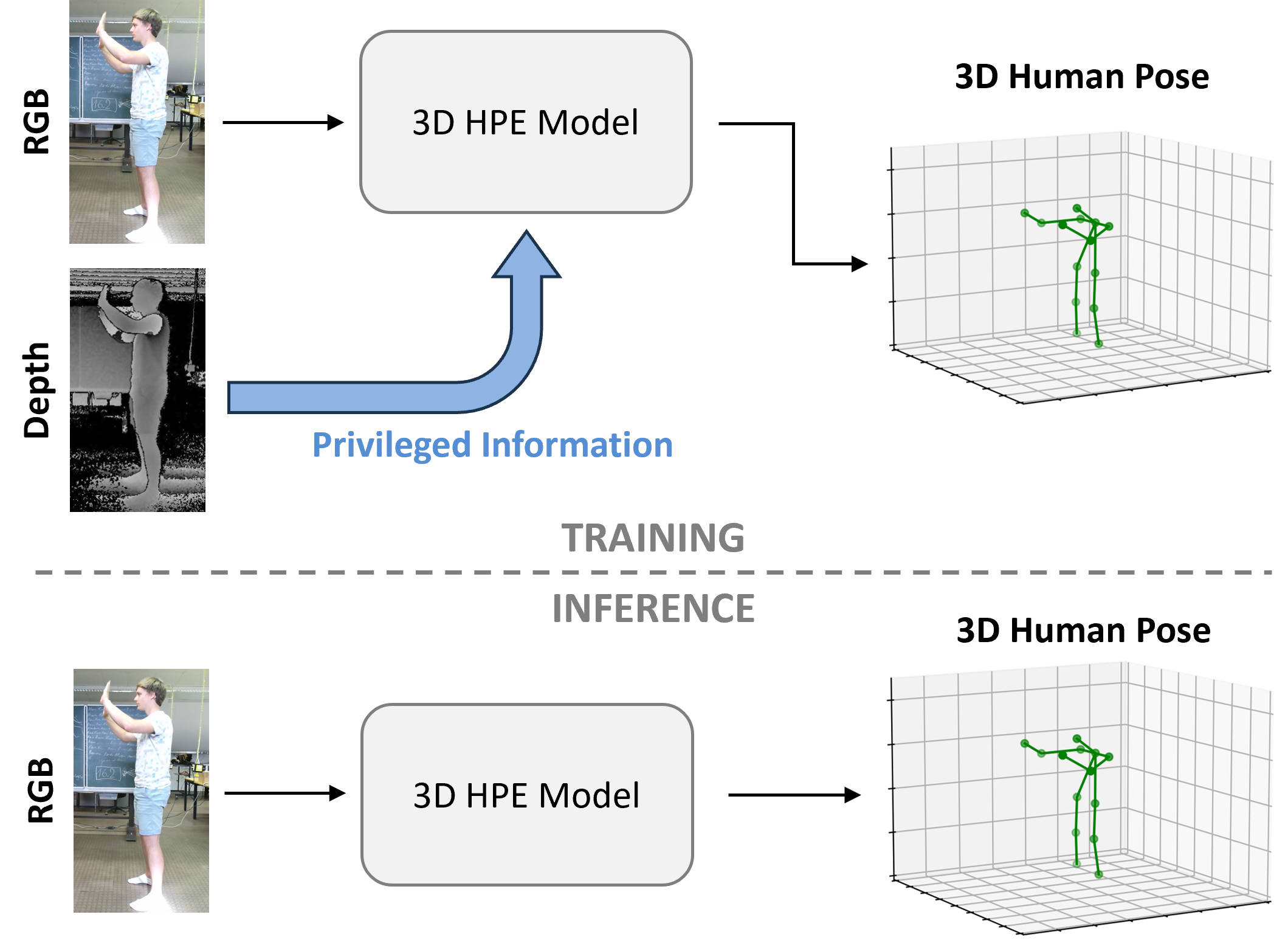}
    \caption{Overview of the proposed 3D human pose estimator (3D HPE) approach. The method is based on the paradigm of Privileged Information~\cite{vapnik2009new}, consisting of providing additional depth data during the training phase. At inference time, the system works only with RGB images.}
    \label{fig:overview}
\end{figure}

The use of depth data could alleviate the depth-ambiguity problem \cite{sarbolandi2015kinect}. Unfortunately, depth cameras are generally limited to indoor environments and less versatile than RGB cameras, which are therefore preferably deployed \cite{mankoff2013kinect}. 
In particular, depth acquisition systems have several drawbacks: stereo cameras are limited by their baseline; multi-camera systems are cumbersome to deploy, require synchronization, and are much more prone to failure; time-of-flight cameras have limited range. Even LiDAR systems are expensive and hard to deploy at scale. Moreover, even if very precise, they may provide a too sparse point cloud to reliably estimate the position of small-volume objects such as people's joints \cite{sarbolandi2015kinect}.

Therefore, in this paper, we propose to leverage the Privileged Information~\cite{vapnik2009new} paradigm, presenting a 3D human pose estimation model that is based only on RGB input images during inference, and that receives side depth information exclusively at training time, as shown in Figure~\ref{fig:overview}.
More specifically, this paradigm enables training an RGB-based hallucination network conditioned by a hallucination loss that forces its feature maps to mimic those produced by the same backbone trained on depth data. 
Thus, the hallucination network learns to extract cues from RGB images that are similar to those that can be extracted from depth images. This eliminates the need for depth input at inference time and simplifies the deployment of the method. 


We apply this paradigm to a heatmap-based method that predicts the 3D pose of humans in world coordinates directly from RGB images.
Specifically, at inference time, the proposed architecture is composed of two input networks, consisting of the effective Small HRNet~\cite{sun2019high} backbone that allows us to keep limited the GPU memory requirements. The extracted feature maps are then concatenated and, through the heatmap-based representation referred as Semi-Perspective Decoupled Heatmaps (SPDH)~\cite{simoni2022semi},  the 3D human pose is finally predicted, as detailed in Figure~\ref{fig:overview}.

In summary, the contributions of this work are the following: 
\begin{itemize}
    \item we apply the paradigm of Privileged Information in the vision-based 3D human pose estimation task; to the best of our knowledge, this is the first investigation in this context, paving the way for future related works;
    \item we prove that the additional depth information positively contributes to the final 3D pose estimation, even if based only on RGB frames at inference time. In other words, we show the possibility of limiting the acquisition of depth data only for the training stage.
\end{itemize}

\section{Related Work}

\subsection{Privileged information}
The first explicit formulation of a method leveraging Privileged Information (also called \textit{Side Information}) was introduced by Vapnik \etal~\cite{vapnik2009new}. The main idea was to leverage additional knowledge only during training in order to improve system performance at the time of testing: this learning method is usually referred to as \textit{Learning Using Privileged Information} (LUPI).
Further works~\cite{xing2002distance,vapnik2009learning,sharmanska2013learning} demonstrated the strength of this approach when multi-modal data is available at least at training time, yielding remarkable improvements in several domains by hallucinating different modalities~\cite{pini2019video}. For instance, Chen \etal \cite{chen2020learning} hallucinated a bird-eye view of an urban scene while driving an autonomous agent, whereas \cite{alehdaghi2022visible} leveraged privileged infrared information for re-identification purposes.
Applied to different architectures, Xu \etal~\cite{xu2015distance} proposed to use depth images to improve distance metric learning for the person re-identification task on RGB images.
A similar approach has been used by Hoffman \etal~\cite{hoffman2016learning}, where the authors trained a CNN for RGB-based object recognition by incorporating insights into the training phase. The hallucination branch, trained on RGB images, is capable of mimicking medium-level features of the depth branch, improving the final accuracy score.
Conversely, Borghi \etal \cite{borghi2018face} proposed to use RGB images as side information in a vision-based system based on depth maps for the face verification task~\cite{borghi2019driver}.

Recently, the paradigm has been adopted for the 3D hand pose estimation task~\cite{yuan20193d}: depth data is used during the training to improve the accuracy of the method which predicts the hand pose from RGB images.  
Finally, the adoption of side information has been applied to the human pose estimation task for enabling the recognition of the correct posture~\cite{lee2023human}. 
Differently from our work, their method focuses on the 2D pose estimation task in low-light scenarios. Here the Privileged Information paradigm is leveraged to disambiguate images with low light intensity.

\subsection{3D Human Pose Estimation}
Recently, the task of single-person 3D pose estimation from single RGB images has become popular. Three main approaches can be identified in the literature. In the first family, methods detect the 2D pose and then project keypoints in the 3D space \cite{lee2018propagating, hossain2018exploiting, mehta2017monocular, martinez2017simple}. In the second case, methods jointly estimate 2D and 3D poses \cite{mehta2017vnect, dabral2018learning, yang20183d, tekin2017learning}. Finally, in the third case, literature methods predict the final 3D pose directly from monocular and single RGB images~\cite{sarandi2018synthetic, sun2018integral, nibali20193d, luvizon20182d}.

The large majority of the literature works belong to the first category, and use publicly available off-the-shelf 2D human pose estimation methods or present a specific module in their pipeline \cite{moreno20173d, chen20173d}.
We observe that these methods can suffer the presence of occlusions, due to which the lifted keypoints can be on the wrong surface and then will project to an inaccurate 3D location.

The other approaches are also not exempt from drawbacks. The joint learning of 2D and 3D poses is often based on large-scale datasets with only 2D pose ground truth annotations, and the final 3D pose is retrieved through structure or anatomical skeleton priors \cite{kanazawa2018end, pavlakos2018learning}. The need for large annotated skeleton datasets can be prohibitive for certain applications, such as pose estimation in constrained or privacy-sensitive work environments. In our work, we aim to develop a system that does not rely on any prior assumption about the skeleton anatomy.
On the other hand, the methods that directly predict the 3D pose from monocular images are based on fine discretizations of the 3D space around the human body, and they usually require a significant computational effort in terms of GPU memory \cite{fabbri2020compressed, pavlakos2017coarse}.
The proposed method leverages a lightweight architecture that performs a 3D pose estimation of the human body directly from RGB images. Furthermore, only at training time, it relies on visual features learned from depth images in order to extract additional cues for a better understanding of the 3D scene captured by the camera.

\begin{figure*}[t!]
    \centering
    \includegraphics[width=1\textwidth]{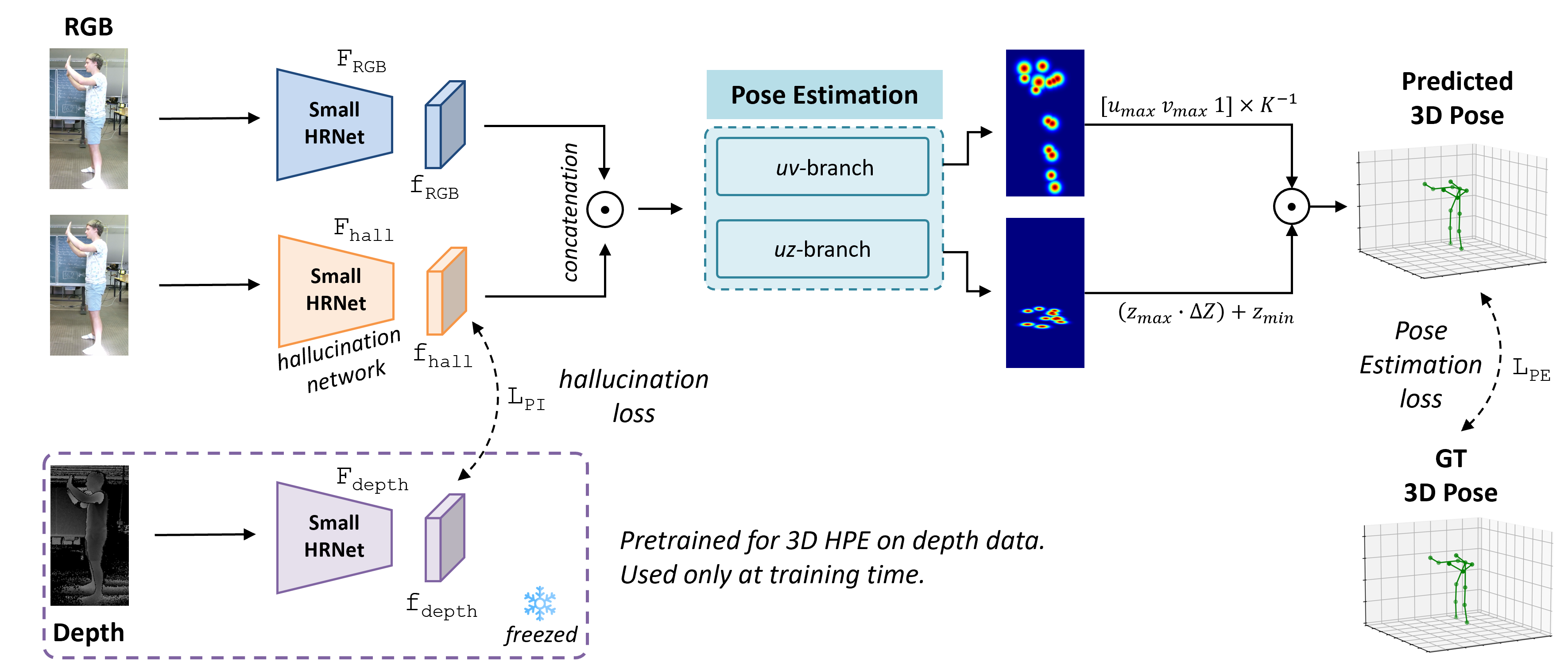}
    \caption{The general overview of the proposed method. 
    Adopting the Privileged Learning \cite{vapnik2009learning} paradigm, RGB and additional depth data are provided during the training stage. In this manner, through a specific loss, we force the hallucination network, to extract features resembling the ones learned by a model pretrained on depth images. These features are concatenated to the visual RGB features and then given as input to the pose estimation branch based on the SPDH \cite{simoni2022semi} representation. Finally, the 3D human pose is predicted in world coordinates.}
    \label{fig:model}
\end{figure*}

\section{Proposed Method}
In this section, we describe in detail the proposed method which leverages depth-based Privileged Information during training and estimates the 3D human pose using an intermediate heatmap-based representation (SPDH)~\cite{simoni2022semi}. 
A general overview of the adopted architecture is depicted in Figure~\ref{fig:model}.

\subsection{Depth-based Privileged Information} \label{sec:privileged}
The paradigm of Privileged Information is applied to the visual encoding step and consists of a two-step training procedure. The network backbone is replicated three times: $F_{depth}$, $F_{RGB}$ e $F_{hall}$. $F_{depth}$ is trained individually on depth data, while $F_{RGB}$ and $F_{hall}$ are trained simultaneously on RGB data only. While $F_{depth}$ and $F_{RGB}$ learn a domain-specific encoding of the scene, $F_{hall}$ is able to capture an intermediate representation between RGB and depth by means of a hallucination loss term. Considering the replication of the visual backbone, we opted for a lightweight version of the well-known HRNet backbone~\cite{sun2019deep}, called SmallHRNet~\cite{sun2019high}.
This operation limits the number of model parameters ($28.5$M vs $3.9$M, for HRNet and SmallHRNet, respectively) and the GPU memory requirements while maintaining good performances in terms of quality and speed. 

The whole training procedure is based on two different and sequential steps, as described in the following.

\subsubsection{Depth-based pretraining.} \label{sec:depth_training}
\begin{figure}[t]
    \centering
     \includegraphics[width=0.95\linewidth]{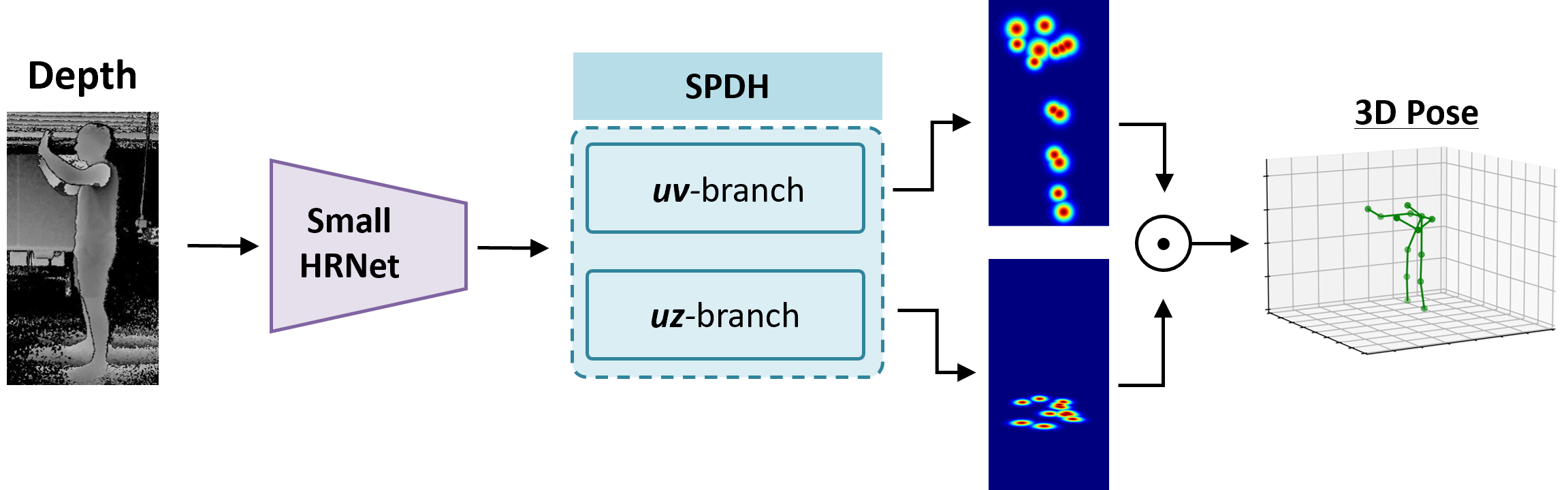}
    \caption{Visual representation of the $F_{depth}$ model (SmallHRNet~\cite{sun2019high} backbone) that predicts the 3D human pose exploiting the SPDH representation. This is the first step of the whole training procedure based on the Privileged Information paradigm.}
    \label{fig:baseline_depth}
\end{figure}

The model $F_{depth}$ (color purple in Figure~\ref{fig:model}) is trained using only depth data as input, without the other two backbones.  
This model is based on the SPDH representation to solve the task of 3D pose estimation. 
The generated embedding $f_{depth}$ are then fed in the Pose Estimation module to decode the SPDH heatmaps.
The training pipeline is visually summarized in Figure~\ref{fig:baseline_depth}.

\subsubsection{RGB training.} 
Once the depth model $F_{depth}$ has been trained, its weights are frozen and two other instances of the backbone are activated (blue and orange components in Figure~\ref{fig:model}). The two networks $F_{RGB}$ and $F_{hall}$ receive as input an RGB image and generate different and complementary embeddings $f_{RGB}$ and $f_{hall}$ thanks to a hallucination loss that forces the hallucination network $F_{hall}$ (orange) to produce intermediate features similar to the ones produced by the backbone $F_{depth}$. During this second training step as well as during inference, the heatmaps are decoded from the concatenation of two embeddings, coming from the two RGB backbones ($F_{RGB}$ and $F_{hall}$).

\subsection{Pose Estimation Branches}
Another key element of the proposed system is the direct estimation of the 3D pose through heatmap generation. In particular, we selected the Semi-Perspective Decoupled Heatmaps (SPDH)~\cite{simoni2022semi} representation which maps each 3D joint location into two decoupled bi-dimensional spaces: the $uv$, \ie the camera image plane, and the $uz$ space, the projection of $z$-values along the $u$ dimension. The two heatmaps are generated from a common embedding obtained concatenating the visual features extracted by the RGB backbone and the hallucination network.

\begin{figure*}[t!]
    \centering
     \includegraphics[width=0.7\textwidth]{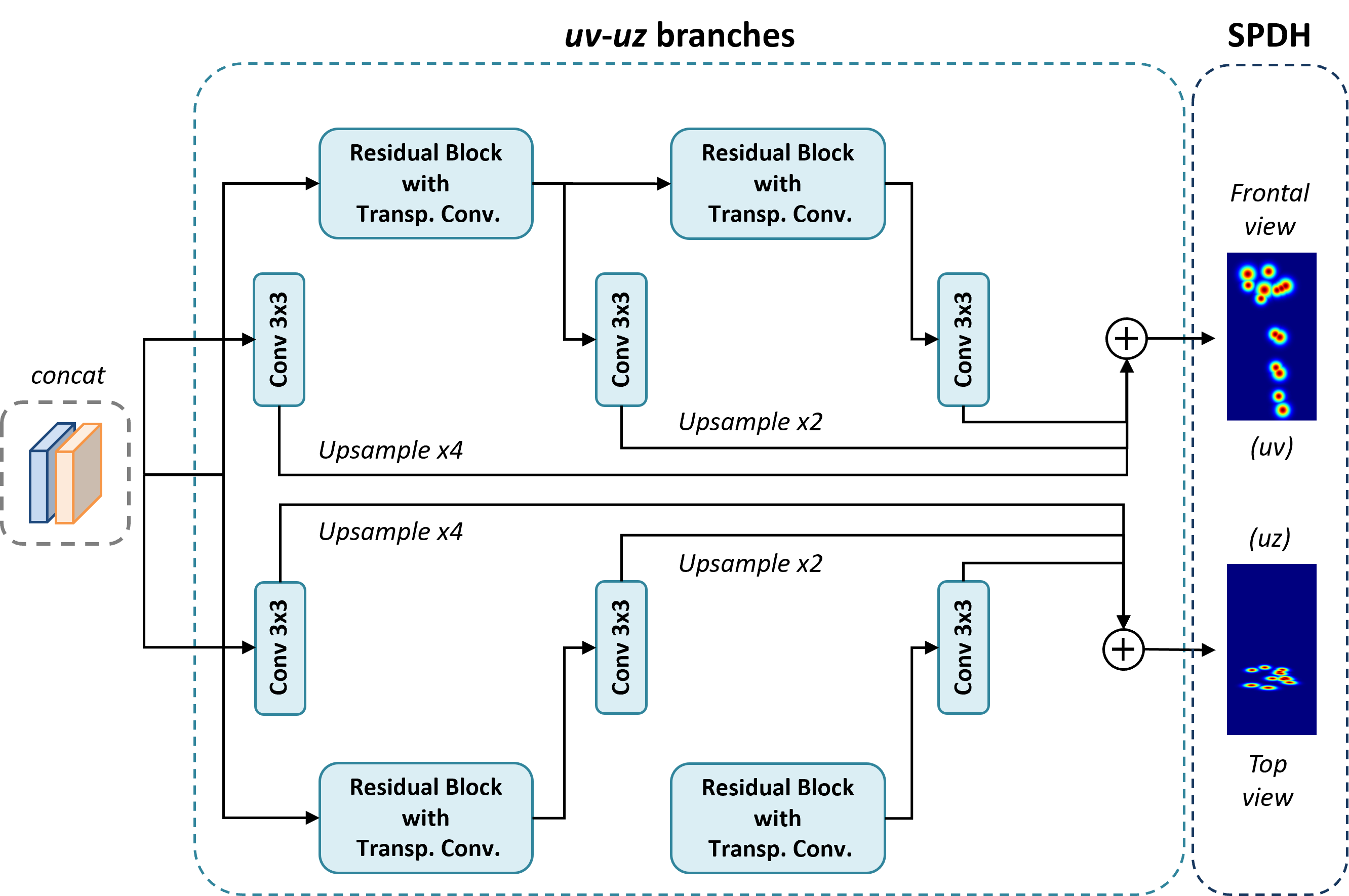}
    \caption{A detailed visualization of the pose estimation branch that leverages the fusion of multi-resolution features to improve the accuracy of the predicted SPDH heatmaps. The input is the concatenation of the feature maps extracted through the hallucination network and the RGB-based one (see Fig.~\ref{fig:overview}).}
    \label{fig:pe_branch}
\end{figure*}

The SPDH output is generated by a double-branch module, as detailed in Figure~\ref{fig:pe_branch}, that leverages multi-resolution features to improve the joint localization precision. In particular, the concatenated features are given as input to a $uv$ and $uz$ branch. Each branch is composed of $3$ convolutional layers with kernel size $3 \times 3$ that predict a set of heatmaps with increasing resolution of $\{\frac{1}{4}, \frac{1}{2}, 1\}$ with respect to the input image size. The upsampling operation is made by a residual block with a transpose convolutional layer. Finally, the multi-resolution outputs are upsampled and summed together in order to obtain the final prediction of dimension $B \times J \times H \times W$.

\subsection{Losses}
The proposed method is trained end-to-end to optimize the SPDH generation leveraging the Privileged Information paradigm through intermediate features learning. We use a standard Mean Squared Error (MSE) loss for the privileged information task and a masked MSE for the pose estimation task:
\begin{equation}\label{eq:loss_pi}
    \mathcal{L}_{PI} = ||f_{depth} - f_{hall}||_2
\end{equation}
\begin{equation}\label{eq:loss_pe}
    \mathcal{L}_{PE} = \frac{1}{|\mathcal{J}|}\sum_{j \in \mathcal{J}} m_j \cdot || H_j - \widehat{H}_j ||_2
\end{equation}

\begin{equation}\label{eq:loss}
    \mathcal{L} = \mathcal{L}_{PI} + \mathcal{L}_{PE}
\end{equation}
where $\mathcal{L}_{PI}$ is the hallucination loss between the visual features extracted from the depth image and the ones extracted from the RGB frame, and $\mathcal{L}_{PE}$ is the pose estimation loss between the predicted heatmap $\widehat{H}_j$ and the ground truth heatmap $H_j$. Moreover, $m_j$ is a visibility mask for each joint and $\mathcal{J}$ is the total number of visible joints considering the $uv$ and $uz$ spaces.

\section{Experimental Validation}
In this section, we explain the experimental settings to evaluate the proposed method, describing the dataset used, the metrics, and the training procedure. Moreover, we present quantitative and qualitative results together with an analysis of the execution time of the system.

\begin{figure*}[t!]
    \centering
    \includegraphics[width=0.95\textwidth]{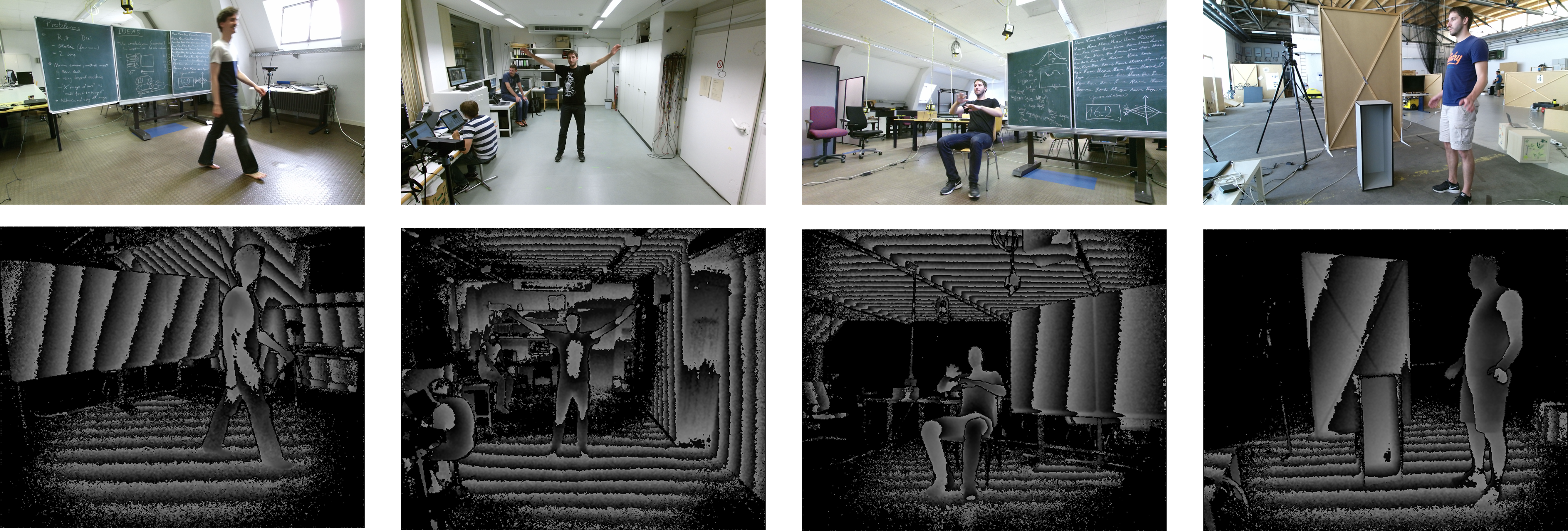}
    \caption{Some visual examples of the Kinect Human Pose Dataset~\cite{zimmermann20183d} in terms of RGB (first row) and depth images (second row). As shown, a variety of body poses, environments, and subject distances, characterize the dataset. Depth maps are depicted in 8-bit format only for visualization.}
    \label{fig:dataset}
\end{figure*}

\subsection{Dataset}
It is important to note that to train and test the proposed system we need a dataset containing RGB and depth input data and 3D annotations of the body joints. 
Unfortunately, only a few datasets are available in the literature, of which we adopt the Kinect Human Pose Dataset~\cite{zimmermann20183d} due to its amount of frames, great variety of poses, and the presence of accurate manual 3D joint annotations.
All available datasets based on the Kinect SDK automatic annotations are not eligible for our task, because of the inaccuracy of the annotations.


\textbf{Kinect Human Pose Datasets~\cite{zimmermann20183d}.} This dataset is composed of two sub-datasets for Human Pose Estimation from RGB-D sensors, namely Multi-View Kinect Dataset (\textit{MKV}) and the Captury Dataset (\textit{CAP}). Each sample provides the RGB image ($1920 \times 1080$ pixels), the depth map ($512 \times 424$ pixels), 3D Human Pose annotation ($14$ keypoints), 3D Kinect SDK Human Pose prediction ($25$ keypoints), and camera calibration parameters. The video sequences consist of $5$ actors ($2$ female and $3$ male), $3$ locations, and $4$ viewpoints with a frame rate of $10$~Hz. The 3D annotations are obtained with a post-processing operation that leverages standard triangulation techniques to lift in the 3D space the 2D poses predicted by an off-the-shelf human pose detector. 
A visual sample of the dataset is reported in Figure~\ref{fig:dataset}.
In our experiments, we focused on the \textit{MKV} dataset. Since our method is influenced by the scene appearance and the original training split contains an important domain shift between the training and the test set, for our experiments we created a more balanced split that we will release together with the code of the proposed method. The new split contains $17360$ training samples (\textit{MKV-t-v2}) and $3886$ test samples (\textit{MKV-e-v2}).


\subsection{Metrics}
To evaluate the accuracy of our 3D pose predictions, we use the average Mean Per Joint Position Error (MPJPE)~\cite{zimmermann20183d} and the mean Average Precision (mAP). The first metric is defined as the mean of the L$_2$ distances between each predicted joint and its corresponding ground truth positions (expressed in centimeters); it conveys the error in the 3D world in terms of translation and rotation, so the lower the better. The latter metric is defined as follows:
\begin{equation} \label{eq:itop_accuracy}
    \text{mAP} = \frac{1}{|N|} \, \sum_{j \in N} \, \big( \lVert \mathbf{y}_j-\widehat{\mathbf{y}}_j \rVert_2 < \tau \big)
\end{equation}
where $N$ is the overall number of joints, $\mathbf{y}_j$ is the predicted joint position while $\widehat{\mathbf{y}}_j $ is the ground truth. This metric represents the accuracy of the MPJPE using different thresholds ($\tau = \{6, 8, 10\}$ centimeters in our experiments) and it improves the understanding of the actual performance of the methods. 

\subsection{Training}
As mentioned in Section~\ref{sec:privileged}, the training procedure is performed in two steps; the first involves a pretraining of the architecture using only depth data as input, while the latter leverages the privileged information for training on RGB images. 
In order to prevent model overfitting, we apply the same data augmentation techniques on both RGB and depth data. In particular, we rotate the images with a range of $[-5, +5]$ degrees, pixel are randomly translated with a maximum range of $15\%$ of the width and $2\%$ of height and horizontal flip. Only during the first step of the whole training procedure, we randomly translate the Z axis of the depth image in the range $[-30cm, +30cm]$.

We use Adam~\cite{kingma2014adam} as an optimizer with a learning rate of $1e^{-3}$ and a decay step with a factor $10$ at the $50\%$ and $75\%$ of the training. Each network is trained for $30$ epochs.

\begin{table*}[t!]
    \centering
    \begin{tabular}{ll ccc ccc c}
        \toprule
        & & & & & \multicolumn{3}{c}{\textbf{mAP} (\%) $\uparrow$} & \\
        \cmidrule{6-8}
        \textbf{Model} & \textbf{Input} & $\mathbf{F_{depth}}$ & $\mathbf{F_{RGB}}$ & $\mathbf{F_{hall}}$ & \textbf{6cm} & \textbf{8cm} & \textbf{10cm} & \textbf{MPJPE} (cm) $\downarrow$ \\
        \midrule
        \textit{SmallHRNet} & \textit{Depth} & \checkmark & & & \textit{50.82} & \textit{61.27} & \textit{68.93} & \textit{11.09} \\
        \midrule
        SmallHRNet & RGB & & \checkmark & & $43.42$ & $52.17$ & $58.61$ & $14.96$ \\
        \midrule
        SmallHRNet & RGB & \textit{frozen} & & \checkmark & $44.43$ & $54.78$ & $60.56$ & $13.90$ \\
        SmallHRNet & RGB & \textit{frozen} & \checkmark & \checkmark & $\mathbf{47.77}$ & $\mathbf{58.13}$ & $\mathbf{65.19}$ & $\mathbf{12.35}$ \\
        \bottomrule
    \end{tabular}
    \caption{Comparison on \textit{MKV-e-v2} between SmallHRNet baselines trained on depth and RGB images and our depth-based privileged information approach trained with and without the RGB-based backbone.}
    \label{tab:MKV_results}
\end{table*}

\subsection{Quantitative results}
Results of the proposed method are reported in Table~\ref{tab:MKV_results}. 
In the first line, we test the SmallHRNet model trained only with depth data, following the step discussed in Section~\ref{sec:depth_training} and visually summarized in Figure~\ref{fig:baseline_depth}. It is important to note that this test represents the upper bound for the performance as we solve the 3D pose estimation task with the same model but using depth images as input data both for training and inference.
In the second line, we report the performance of the method using only the model trained on RGB images. These results are then used as a reference to quantify the boost of performances achieved by our method thanks to the introduction of Privileged Information in an RGB-based system. 
We tested our Privileged Information approach in two configurations: the first one uses only the hallucination network, while the latter leverages both the hallucinated features and the embedding representing the RGB domain. As can be seen in Table~\ref{tab:MKV_results}, the Privileged Information paradigm improves the performance even with just the hallucination network with respect to the RGB-only baseline. When adding the domain-specific features of the RGB images, the system improves even more obtaining a $-2.66$ in terms of MPJPE and $+4.5\%$ in terms of mAP with a threshold $\tau=10$cm.

\begin{figure*}[t!]
    \centering
    \includegraphics[width=1\textwidth]{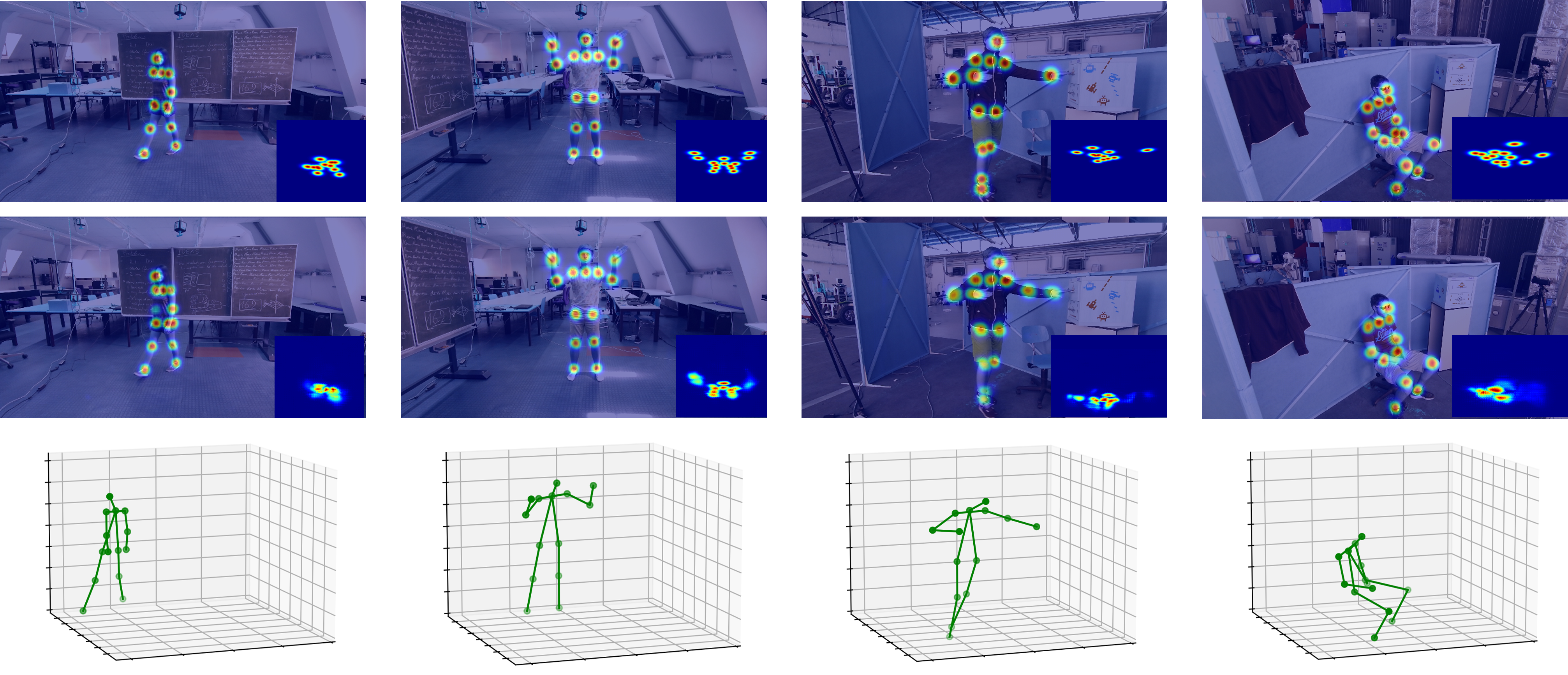}
    \caption{Sample outputs of the proposed method. The ground truth and the predicted uv-heatmaps are superimposed on the input RGB frame in the first (ground truth) and second (prediction) lines, respectively. In the bottom right corners are depicted the uz-heatmaps. In the third line, it is represented the predicted 3D human pose. As shown, our method is able to correctly handle a variety of poses and backgrounds, producing plausible final poses. }
    \label{fig:final}
\end{figure*}

\subsection{Qualitative Analysis}
To provide better insights of the capabilities of the proposed approach, we provide some qualitative results.
A few samples of the predictions of our approach are depicted in Figure~\ref{fig:final}, showing a comparison between the ground truth and the predicted poses in terms of SPDH representation and 3D skeleton.
It can be seen how the model is able to accurately estimate the position of the joints in the image plane, as well as the pose of the humans in the 3D space.

\begin{figure}[t]
    \centering
    \includegraphics[width=0.6\columnwidth]{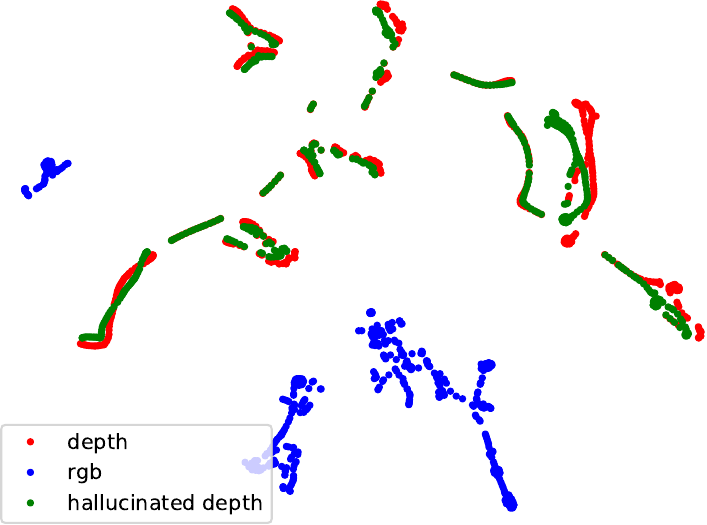}
    \caption{T-SNE of features extracted from RGB data and depth data, as well as hallucinated depth features generated from RGB data. For each of the three modalities, we extracted features from 500 samples. Hallucinated depth features are projected in close proximity to the original depth features, demonstrating the effectiveness of the method.}
    \label{fig:tsne}
\end{figure}

We then perform an additional experiment, projecting features extracted from our model with T-SNE~\cite{van2008visualizing}. We sample 500 frames from the dataset and extract the corresponding features using the three branches in our model, $F_{depth}$, $F_{RGB}$ e $F_{hall}$. The aim of this experiment is to qualitatively inspect whether the learned hallucinated features are indeed close to the real features that we would extract when having depth data available.
Interestingly, the 2D T-SNE projection shown in Figure~\ref{fig:tsne}, clearly highlights the proximity of the features obtained from depth data and the hallucinated ones inferred from RGB data. This further confirms the effectiveness of our approach.

\subsection{Execution Time Analysis}
In the final part of the experimental analysis, we focus on the investigation of speed execution.
We compute the inference on a computer equipped with an Intel Core i7-7700K (4.50GHz) and an Nvidia GeForce GTX 1080 Ti, obtaining $33$ frame per second with a video memory occupation of about $2.0$ GB. 
These values assure real-time performance of the proposed system and enable its use in devices with limited computational power. 

\section{Conclusion and Future Work}
In this paper, we presented a method for 3D human pose estimation from RGB images based on the Privileged Information paradigm. 
In particular, we showed how it is possible to use depth data to improve the performance of a system based only on RGB images at inference time. We proved that the Privileged Information paradigm is effective even when applied to limited and small datasets. Moreover, the paradigm is general and can be applied to boost the performance of any method for 3D human pose estimation. In this paper, we applied it to a method that represents 3D poses as semi-perspective decoupled heatmaps and observed a significant improvement in performance.
Possible future works include collecting a larger dataset with accurate 3D joint annotations in order to broaden the experimental validation, and testing the paradigm using estimated depth from RGB (\eg,~\cite{ranftl2020towards, ranftl2021vision, mildenhall2021nerf, kim2022global, oquab2023dinov2}), which would remove the need for depth data completely. 


\section*{Acknowledgement}
This work was partially supported by the Piano per lo Sviluppo della Ricerca (PSR 2023) of the University of Siena - project FEATHER: Forecasting and Estimation of Actions and Trajectories for Human-robot intERactions.

This work was supported by the European Commission under European Horizon 2020 Programme, grant number 951911—AI4Media.

Part of the work was funded within the project ``AI platform with digital twins of interacting robots and people'', FAR Dipartimentale 2022 DIEF - Unimore.

%
%
\bibliographystyle{splncs04}
\bibliography{ref}
\end{document}